\documentclass[10pt,twocolumn,letterpaper]{article}

\usepackage{wacv}
\usepackage{times}
\usepackage{epsfig}
\usepackage{graphicx}
\usepackage{amsmath}
\usepackage{amssymb}
\usepackage{multirow}
\usepackage{arydshln}
\usepackage{colortbl}
\usepackage{color}
\usepackage{float}
\usepackage{caption}

\wacvfinalcopy 

\def\wacvPaperID{1037} 

\ifwacvfinal\fi
\begin{document}

\title{Semantic Consistency and Identity  Mapping Multi-Component Generative Adversarial Network for Person Re-Identification}

\author{Amena Khatun \hspace{2cm} Simon Denman \hspace{2cm} Sridha Sridharan \hspace{2cm} Clinton Fookes \\
Image and Video Laboratory,
Queensland University of Technology (QUT), Brisbane, QLD, Australia\\
{\tt\small Email: \{a2.khatun, s.denman, 
s.sridharan, c.fookes\}@qut.edu.au}
}

\maketitle
\ifwacvfinal\thispagestyle{empty}\fi

\begin{abstract}
In a real world environment, person re-identification (Re-ID) is a challenging task due to variations in lighting conditions, viewing angles, pose and occlusions. Despite recent performance gains, current person Re-ID algorithms still suffer heavily when encountering these variations. To address this problem, we propose a semantic consistency and identity mapping multi-component generative adversarial network (SC-IMGAN) which provides style adaptation from one to many domains. To ensure that transformed images are as realistic as possible, we propose novel identity mapping and semantic consistency losses to maintain identity across the diverse domains. For the Re-ID task, we propose a joint verification-identification quartet network which is trained with generated and real images, followed by an effective quartet loss for verification. Our proposed method outperforms state-of-the-art techniques on six challenging person Re-ID datasets: CUHK01, CUHK03, VIPeR, PRID2011, iLIDS and Market-1501.
\end{abstract}
\vspace{-2mm}
\vspace{-2mm}
\section{Introduction}
\vspace{-2mm}
\begin{figure}
\begin{center}
\includegraphics[width=0.9\linewidth]{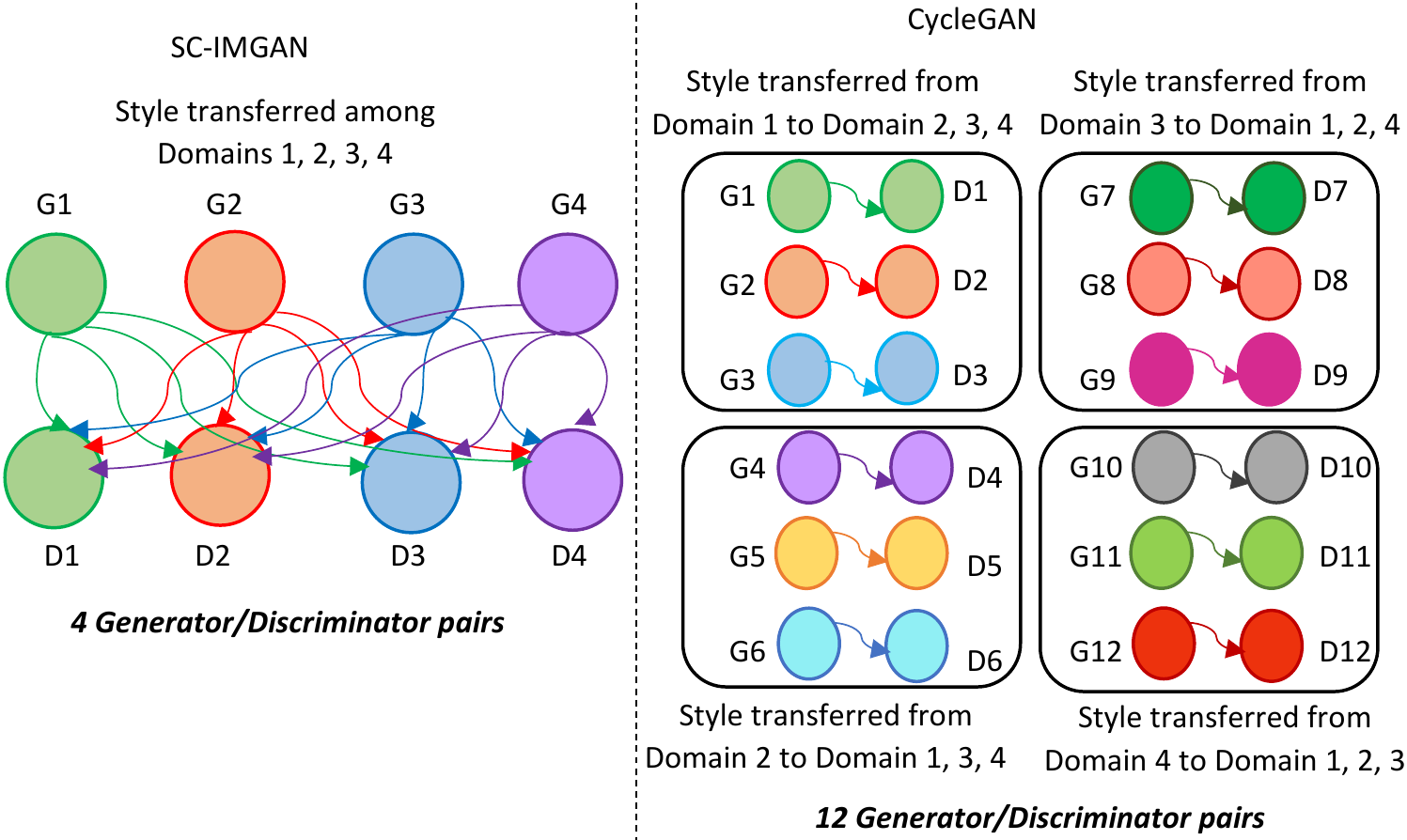}
\end{center}
\caption{Illustration of the proposed SC-IMGAN and the existing CycleGAN \cite{CycleGAN2017}. CycleGAN translates images between two domains at a time; while SC-IMGAN can transfer images between multiple domains with an identity mapping loss to ensure that images retain their identity after translation, and a semantic consistency loss to preserve semantic information shared across domains. Thus for four domains, CycleGAN requires 12 generator/discriminator pairs while SC-IMGAN requires only 4 pairs.}
\label{fig:comparison}
\vspace{-4mm}
\end{figure}

Person re-identification (Re-ID) aims to match an image of a person to a large gallery set, where probe and gallery images are from different cameras. Although person Re-ID is a widely investigated research area, it is still challenging to re-identify the target person accurately in the presence of domain variations including changes due to illumination, pose, viewing angle, and background. Thus in a real world scenario where the domain of the target images has no overlap with the gallery images, performance is severely reduced. To address the domain variation challenge, previous researchers adopted feature extraction methods to learn discriminative features across different cameras. However, while these methods have helped relax the closed-world assumptions of previous methods, performance is degraded when confronted with a real-world scenario where target image conditions are unseen. 

Motivated by this problem, we propose a multi-component generative adversarial network for style adaptation, from one to many domains, to improve the discriminative ability of a CNN trained for person Re-ID. Specifically, for image domain translation, a multi-component model is proposed to generate synthetic images where the style of a person is transferred from one domain to multiple other domains with an identity mapping loss and a semantic consistency loss. The identity mapping loss is used to ensure that the identity of the transferred person is the same as the original person, and the semantic consistency loss is used to encourage the network to preserve the learned embedding during domain translation. Thus the Re-ID model will be trained with a larger set consisting of real and synthetic images for the same person showing different styles, such as changes in background and lighting.

Recently, CycleGAN \cite{CycleGAN2017} has been used by person Re-ID researchers to transfer style from camera-to-camera or domain-to-domain, however CycleGAN can only transfer images between two domains at once. In contrast, our proposed SC-IMGAN is able to generate new images with the style of multiple domains at a time, all with the same identity as the original image. As shown in Figure \ref{fig:comparison}, let us assume we have 4 domains. To transfer styles among these 4 domains, CycleGAN requires 12 generator/discriminator pairs whereas the proposed SC-IMGAN requires only 4 generator/discriminator pairs. Our work also differs in that the generator of SC-IMGAN aims to keep the same identity after translation between domains. CycleGAN uses only the cycle-consistency loss between the real and the reconstructed images at the pixel-level, and as such  fails to capture semantic features shared across domains. This leads to a drop in performance when domains are vastly different. To address this limitation, we propose a semantic-consistency loss operating at the feature-level (i.e. on the embeddings learnt by the encoder) to ensure consistent semantic information is extracted for both the input and style transferred images. These newly generated unlabeled synthetic samples are then used as additional positive images to train the Re-ID network alongside real images. Hence, the trained network benefits as it learns different appearance variations (pose, lighting and background) for a  person.

Within the proposed framework, we adopt the verification-identification approach of \cite{Amena}, with a quartet loss. However, the quartet loss of \cite{Amena} does not specify how close the intra-class features should be in feature space, resulting in a drop in performance due to large intra-class distances between images of the same identity. As such, we propose an improved quartet loss which forces the network to minimise the intra-class distance more than the inter-class distance, regardless of whether the positive and negative pairs share the same probe image or not, and simultaneously ensures intra-class features are close to each other, improving network generalisation. Without this, images of the same class may form a large cluster with a relatively large average intra-class distance in feature space which is not desirable. The contributions of this paper are:
\begin{itemize}
    \item We propose to generate synthetic images with a multi-component generative adversarial network where the images from one domain are transferred to all other available domains simultaneously.
    \vspace{-2mm}
    \item We demonstrate how identity can be better preserved during style transfer by using an identity mapping loss and a novel semantic consistency loss.
    \vspace{-2mm}
    \item We propose a novel improved quartet loss to minimise the distance between images of the same identity more than the distance between dissimilar identities, improving the generalisation of the Re-ID network.
    \vspace{-2mm}
    \item We exceed the state-of-the-art accuracy compared to existing methods on six challenging person Re-ID datasets: CUHK01 \cite{6619305}, CUHK03 \cite{6909421}, VIPeR \cite{1478416}, PRID2011 \cite{conf/scia/HirzerBRB11}, iLIDS \cite{Wang2014PersonRB} and Maerket-1501 \cite{Market}.
\end{itemize}
\vspace{-2mm}
\section{Related Work}
\vspace{-2mm}
In this section, we briefly summarise related research in image domain translation using GANs, and deep learning based person Re-ID.
\vspace{-2mm}
\subsection{Image Domain Translation by Generative Adversarial Networks (GANs)}
\vspace{-2mm}
Generating realistic synthetic images from real images is a challenging task, and requires a model to capture the distribution of the real images. To generate synthetic images which have similar properties to the original training dataset, GANs were introduced in \cite{NIPS2014_5423}. Generally, GANs use noise to synthesise an image and the network is trained in an adversarial manner. Inspired by the success of GANs, various extensions have been proposed for image-to-image translation \cite{7780634,8658643}, pixel-level transfer from source to target domains \cite{Yoo2016PixelLevelDT}, and style transfer between domains \cite{Yi2017DualGANUD}. Rather than using noise alone as the stimulus, the conditional GAN (cGAN) \cite{8100115} is proposed to control the mode of generated images, however, cGANs need a pair of images for training which is not available for many tasks. To address this, \cite{NIPS2016_6544} introduced coupled GANs which use a pair of GANs instead of a pair of images. CycleGAN \cite{CycleGAN2017} also overcomes the requirement of paired data through a cycle consistency model, where the source domain image is translated according to the target domain and vice-versa. However, CycleGAN can transfer the styles only between two domains at a time. Similar to CycleGAN, \cite{Yi2017DualGANUD} proposed DualGAN for unpaired image-to-image translation using dual learning to train the translator network with two sets of images from two domains. Although in StarGAN \cite{8579014}, a multi-domain translation network is proposed using a single generator which takes one-hot vector along with each input to represent domain information, this method is only applied when there is no feature mismatch between domains such as face attribute modification, where all the domains have slight shifts in qualities of the same category of images: human faces with a clear background. Moreover, the restrictive nature of modeling all mapping function as a single network may create problems when the mapping functions between different pairs of domains varies.

As the performance of person Re-ID drops severely due to variations between domains or cameras, researchers adopted GANs for image translation to generate synthetic images with different styles so that CNNs can be trained with multiple styles of a person. Re-ID researchers \cite{zheng2017unlabeled,zhong2018camera,Wei2018PersonTG} have typically adopted the traditional GAN \cite{NIPS2014_5423} or CycleGAN \cite{CycleGAN2017} to generate synthetic images which are used to train a Re-ID network. In \cite{zheng2017unlabeled}, the GAN is first introduced for person Re-ID to generate new samples which are not present in the training data. However, \cite{zheng2017unlabeled} only generated new samples for data augmentation instead of increasing the number of positive pairs. To translate the images between two domains, \cite{zhong2018camera,Wei2018PersonTG,8485427} employed CycleGAN, aiming to find a mapping function between two domains. Although these methods achieve promising performance, they don't consider moving beyond two domains. As most camera networks contain 10's or even 100's of cameras, the ability to transfer between an arbitrary number of domains is required. As preserving person identity is crucial for Re-ID, \cite{Wei2018PersonTG, Bak_2018_ECCV} propose adding an identity preserving loss using a foreground mask, however, they require an additional network and extra supervision to extract the mask images.

\vspace{-2mm}
\subsection{Deep Learning For Person Re-ID}
As deep CNN (DCNNs) combine feature extraction and metric learning \cite{8099654, 8100017} in a single framework, person Re-ID researchers adopted DCNNs to achieve state-of-the-art performance. Siamese networks are adopted by \cite{DBLP:journals/corr/VariorHW16,7780513,6909421}, and a pair of images are taken as input and the network is trained to push images of the same identity close to each other in feature space. Other researchers adopted a triplet network which minimises the intra-class distance with respect to the same probe image. \cite{Cheng_2016_CVPR} improved the original triplet loss by adding new constraints to further minimise intra-class distance using a second margin; and \cite{DBLP:journals/corr/ChenCZH17} further improved the triplet loss through the use of two margins: the first performing the same function as the original triplet loss, while the second seeks to maximise inter-class distance. However, the second margin is weaker than the first, which leads to the network being dominated by the triplet loss,
i.e. minimising the intra-class distance when the probe images
belong to the same person. To address this problem, a new loss is proposed in \cite{Amena} to force the network to minimise the intra-class distance more than the inter-class distance, regardless of whether the probe image comes from the same person or not. However, \cite{Amena} does not specify how close the positive pair should be in feature space. In contrast to \cite{DBLP:journals/corr/ChenCZH17,Amena}, we propose a new loss function for Re-ID which not only reduces the intra-class distance more than the inter-class distance with respect to multiple different probe images, but also specifies how close the positive pair should be in feature space.

A number of recent approaches have used part based methods \cite{Li_2017_CVPR,zhao2017spindle,98a1e05749b24099a51dcf3c22daefd9, Yang2019PatchBasedDF} with the aim of overcoming occlusions, and handling partial observations. These methods, however, all rely on verification or identification frameworks only instead of jointly adopting both approaches, and some methods require additional supervision and pose annotation. Other methods \cite{Fan:2018:UPR:3282485.3243316, DECAMEL, Lin2019ABC} are focused on clustering or transfering the knowledge from a labeled source dataset to an unlabelled data using pseudo labels. However, different identities may have the same pseudo label which can make it hard for the model to distinguish similar people.

Our work differs from the above approaches in architecture, loss function and motivation. Rather than only address the issue of image translation between two domains, we propose a multi-component adversarial network to transfer the style from one to many domains at once. To improve person Re-ID performance, we add an identity mapping loss to preserve the identity of transferred images. In addition to the cycle consistency loss applied at pixel level, we propose a semantic-consistency loss applied at the feature-level to capture shared semantic content for flexible cross-domain image translation. The real and style transferred images are then fed into the proposed four-stream CNN with the improved quartet loss for verification, and softmax loss for identification. Finally, as in a real world situation gallery and query images likely have no overlap, we build a cross-domain architecture to cope with such a scenario.

\vspace{-2mm}
\section{Proposed Method}
\begin{figure*}
\begin{center}
\includegraphics[width=1.0\linewidth]{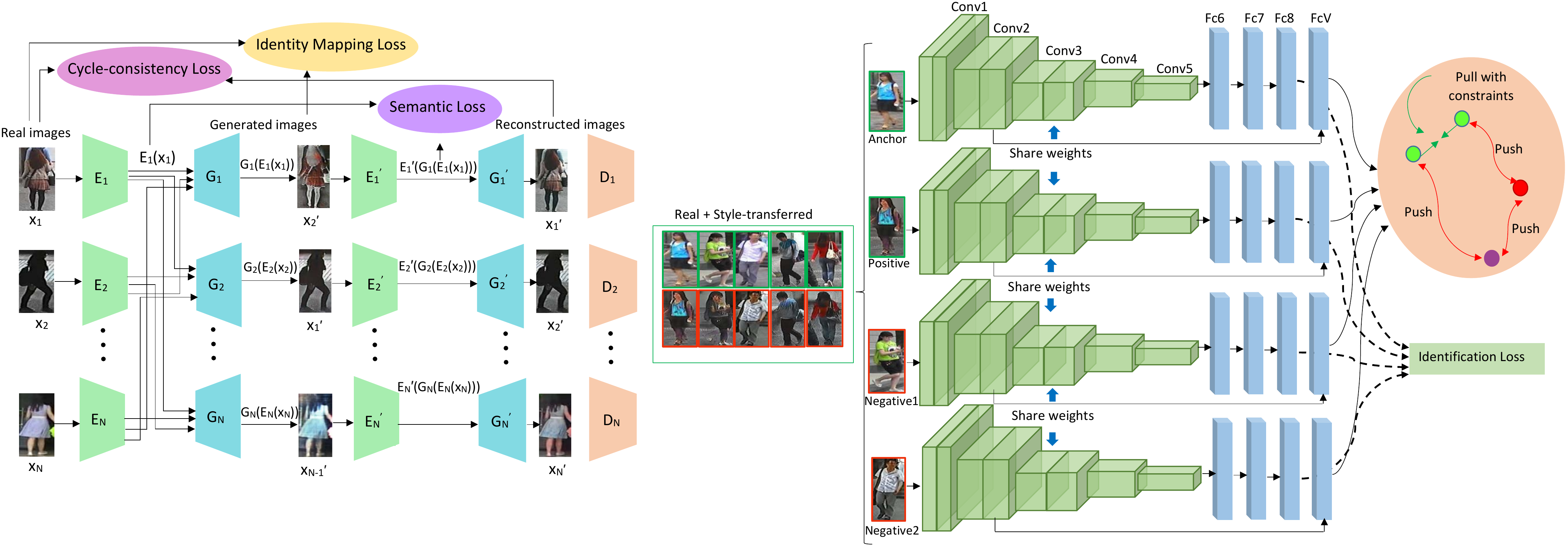}
\end{center}
\caption{Architecture of the proposed model. At first, style transferred images are generated by SC-IMGAN. The cycle-consistency loss and identity mapping loss are applied at the pixel level while the semantic-consistency loss is appiled at the feature level. We use six domains (CUHK01, CUHK03, VIPeR, PRID2011, iLIDS and Market-1501) to transfer the styles of pedestrians. The style transferred images are then concatenated with real images to train the proposed four-stream deep CNN model. The extracted features are fed into the verification and identification losses to identify the person.}
\label{fig:architecture}
\vspace{-4mm}
\end{figure*}

\vspace{-2mm}
The proposed architecture is shown in Figure \ref{fig:architecture}, and consists of two networks: one for image domain translation; and one for Re-ID. These networks are explained in the following subsections.
\vspace{-2mm}
\subsection{Semantic Consistency and Identity Mapping Multi-Component GAN}
\vspace{-2mm}
The widely adopted CycleGAN learns to map between only two domains at a time. We propose an identity mapping and semantic feature preserving multi-component adversarial network to address the problem of mapping images when more than two domains exist. 
Let us assume that we have $N$ source domains: $S^1, S^2, S^3,.....S^N$. Thus the proposed method learns to find a mapping among all available domains. For $N$ domains, the proposed method requires $N$ generators and discriminators where each of the generators contains an encoder and decoder. To compare the distribution of the generated images to the distribution of other domains, adversarial losses are used. For example, if we want to transfer the style of domain $S^1$ $\rightarrow$ $S^2$, the adversarial loss is given by:
\vspace{-2mm}
\begin{align}
L_{GAN}(E_1,G_1, D_2, S^1, S^2) = _{x_2 \sim P_{S^2}} [log D_2(x_2)]  \nonumber	\\
+ _{x_1 \sim P_{S^1}} [log(1-D_2(G_1(E_1(x_1))))],
\label{eq:1}
\end{align}
where the mapping function is $G_1(E_1): S^1$ $\rightarrow$ $S^2$ and $D_2$ is the discriminator. The generator attempts to generate images with the distribution of the new domain ($S^2$) and the discriminator $D_2$ tries to differentiate generated images from real images. However adversarial training requires paired training data, otherwise infinitely many mappings will induce the same distribution over the output; and thus many input images will map to the same output image in the absence of paired training data. To address this problem, we adopt the cycle consistency loss \cite{CycleGAN2017} to translate images from domain $S^1$ to domain $S^2$, and then translate it back from domain $S^2$ to domain $S^1$, and as such do not require paired training data. For example, two domains require two mapping functions which should be bijective. The cycle consistency loss can be expressed as, 
\vspace{-2mm}
\begin{align}
L_{cyc} = _{x_1 \sim P_{S^1}} [ || {G_1}^{\prime}({E_1}^{\prime}(G_1(E_1(x_1)))) - x_1||_1] \nonumber\\+ _{x_2 \sim P_{S^2}} [||{G_2}^{\prime}({E_2}^{\prime}(G_2(E_2(x_2))))- x_2||_1],
\label{eq:2}
\end{align}
where ${G_1}^{\prime}({E_1}^{\prime}(G_1(E_1(x_1))))$ represents the reconstructed version of the real image $x_1$, and ${G_2}^{\prime}({E_2}^{\prime}(G_2(E_2(x_2))))$ is the reconstructed version of the real image, $x_2$ .

As preserving person identity is crucial for person Re-ID, we add an identity mapping loss \cite{taigman2016unsupervised} alongside the cycle consistency loss to force the generator to preserve the identity of the source domain's real images in the target domain, i.e. we require that if the source domain shows an image of person $p$, then the person $p$ is also rendered in the target domain. The identity preserving loss can be expressed as,
\vspace{-2mm}
\begin{align}
L_{identity}= _{x_1 \sim P_{S^1}} [ || G_1(E_1(x_1)) - x_1||_1] \nonumber	\\
+ _{x_2 \sim P_{S^2}} [||G_2(E_2(x_2)) - x_2||_1],
\label{eq:3}
\end{align}
where $ G_1(E_1(x_1))$ and $G_2(E_2(x_2))$ are the style transferred images from the real images, $x_1$ and $x_2$ respectively.

Further, we propose a feature-level semantic-consistency loss to preserve semantics during cross-domain translation, helping maintain identity between vastly different domains. The semantic-consistency loss is given by,
\vspace{-2mm}
\begin{align}
L_{semantic} = _{x_1\! \sim P_{S^1}}\! [ || {E_1}\!^{\prime}\!(G_1\!(E_1\!(x_1)))\! - E_1\!(x_1)||_1] \!\nonumber\\
+ _{x_2 \!\sim P_{S^2}} [||{E_2}\!^{\prime}\!(G_2(E_2(x_2)))\! - E_2(x_2)||_1].
\label{eq:4}
\end{align}
This loss is applied to the embeddings so that the encoder extracts the same high level features for both the input and output, such that semantic information is consistent across domains. Here, ${E_1}^{\prime}(G_1(E_1(x_1)))$ represents the embedding of the translated images from domain $S_1$ to $S_2$ and $E_1(x_1)$ is the embedding of the $S_1$ domain's real images, and similar for the reverse mapping. Thus to transfer the style from domain $S^1$ to $S^2$ with the preserved identity and semantic features, the objective of SC-IMGAN is,
\vspace{-2mm}
\begin{align}
L_{SC-IMGAN} =  L_{GAN}(E_1, G_1, D_2, S^1, S^2) \nonumber \\
+ L_{GAN}(E_2, G_2, D_1, S^2,S^1)  \nonumber \\ + {\lambda_1 L_{cyc}}  +  {\lambda_2 L_{identity}} + {\lambda_3 L_{semantic}}.
\label{eq:5}
\end{align}
When training the proposed multi-component network, we train the mapping between two domains at a time, and iterate through pairs of domains to train all mappings, with the aim of preserving semantic information and person identity. In this work, we consider six domains (CUHK01, CUHK03, VIPeR, PRID2011, iLIDS and Market-1501) for image translation which requires 6 generator/discriminator pairs. Each of the generators are disengaged to utilize half as encoders and the other half as decoders. Thus for each domain, an encoder and a decoder from different domains can be combined to reduce the required number of generators. As such, the network can translate an image from one domain to all other domains.
\vspace{-2mm}
\subsection{Joint Verification-Identification for Person Re-Identification}
\vspace{-2mm}
The newly generated style transferred images are used as the input alongside real images in a CNN for person Re-ID. Here, a four-stream DCNN is proposed to combine verification and identification tasks in a single framework. We propose an improved quartet loss for verification which requires four input images denoted as, $I_i$ = $I_i^1$, $I_i^2$, $I_i^3$, $I_i^4$ where $I_i^1$ is the anchor image, $I_i^2$ is the positive image, and $I_i^3$ and $I_i^4$ are two different negative images. Although great success has been had with the triplet loss for person Re-ID, it suffers from poor generalisation in real world scenarios due to totally unseen target data. The triplet loss pushes images of the same identity close to each other only when the probe images come from the same identity, which is not practical in the real world. We also notice that neither the triplet or quartet loss specify how close the positive pair should be in feature space. Thus intra-class variation within feature space may be high, resulting in a severe drop in performance. To overcome these challenges, we propose an improved quartet loss to minimise the intra-class variation over the inter-class variation, regardless of whether the probe image belongs to the same person or not. We further insert a new term to push the network to minimise the intra-class distance more than the inter-class distance, and ensure this distance is less than a second margin. The full objective of the proposed verification loss is given by,
\vspace{-2mm}
\begin{align}
L_{ImpQuartet} = \sum_{i=1}^{n} \Big(max\big\{\parallel\Theta_w (I_i^1) - \Theta_w (I_i^2) \parallel^2  \nonumber\\ - \parallel \Theta_w (I_i^1) -  \Theta_w (I_i^3) \parallel^2  + \parallel\Theta_w (I_i^1) - \Theta_w (I_i^2)\parallel^2 \nonumber\\ - \parallel\Theta_w (I_i^4) - \Theta_w (I_i^3)  \parallel^2, \tau_1 \big\} \nonumber\\ + max\big\{\parallel\Theta_w (I_i^1) -\Theta_w (I_i^2) \parallel^2, \tau_2 \big\}\Big).
\label{eq:6}
\end{align}
Here, the positive pair, comprised of $\Theta_w (I_i^1)$ and $\Theta_w (I_i^2)$, is included twice in the first term of Equation \ref{eq:6} to compensate for having two negative pairs ($\Theta_w (I_i^1)$, $\Theta_w (I_i^3)$; and $\Theta_w (I_i^4)$, $\Theta_w (I_i^3)$). The first negative pair shares a common probe image with the positive pair (i.e. $\Theta_w (I_i^1)$), while the second negative pair uses two different images. Thus, the proposed loss forces the network to maximize the inter-class distance even if the target image comes from a different identity, while the second term forces the distance between $\Theta_w (I_i^1)$ and $\Theta_w (I_i^2)$ to be less than $\tau_2$, where $\tau_2$ is less than $\tau_1$, ensuring that features for the same identity are close in feature space. Thus by using the improved quartet loss, the inter-class distance is required to be larger than the intra-class distance irrespective of whether the probe image comes from the same identity or multiple different identities; while ensuring that the intra-class features will lie close to each other in the feature space. 
\begin{figure}
\begin{center}
\includegraphics[width=1.0\linewidth]{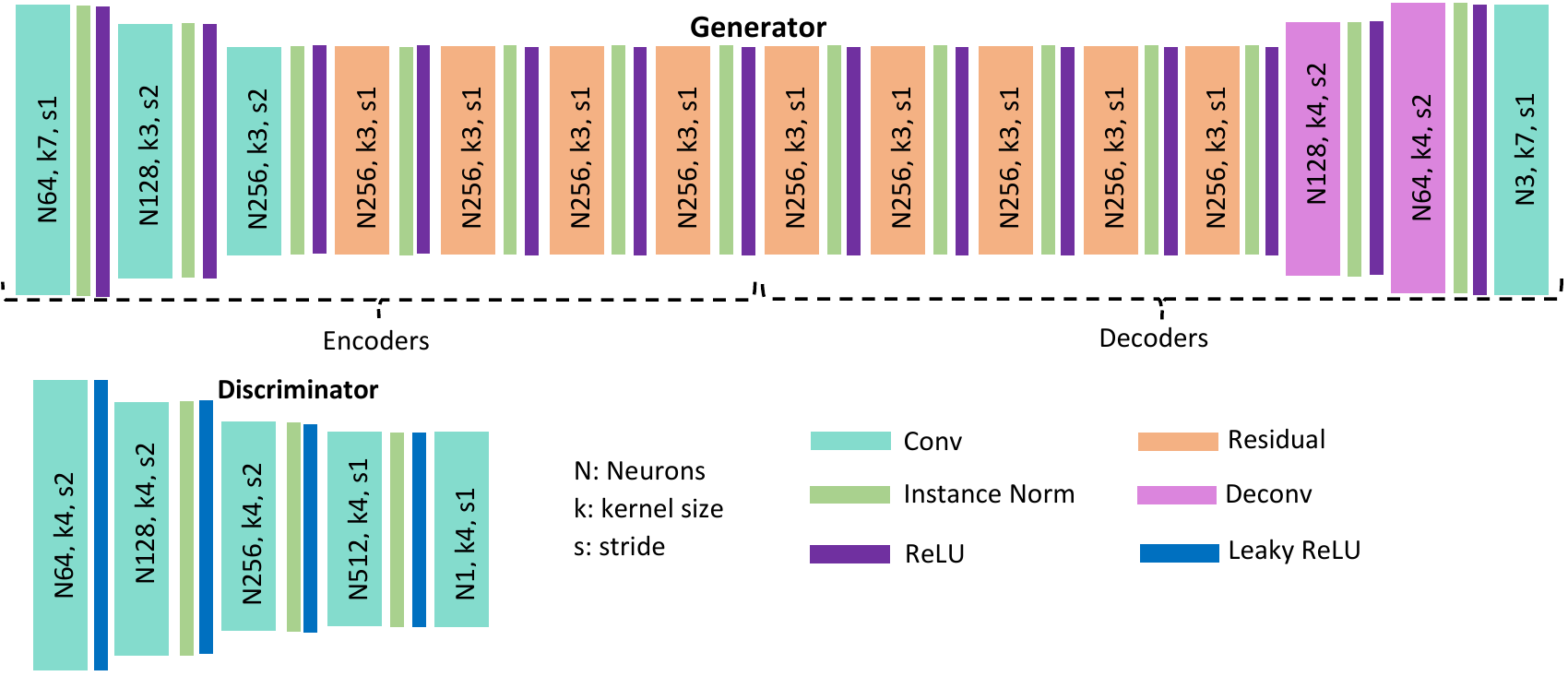}
\end{center}
\caption{Architecture of the GAN where $N$, $k$, and $s$ represents the number of neurons, kernel size and stride respectively. For six domains, our method requires six generator/discriminator pairs, i.e. an encoder, a decoder and a discriminator per domain.}
\label{fig:GAN}
\vspace{-4mm}
\end{figure}


The identification model in this work is the same architecture as \cite{Amena, AAAI1714313} which forces the network to push images of different identities away from each other to identify the query image. As the proposed model trains the network with four input images, three pairs are created for identification, one is a positive pair and the remaining two are negative pairs. For identification, a softmax layer is used to obtain the similarity between the probe and the gallery images. The identification loss is,
\vspace{-2mm}
\begin{equation}
L_{identification}= - \sum_{i=1}^{n}p_ilog\bar{p_i}=-log\bar{p}_t,
\label{eq:7}
\vspace{-2mm}
\end{equation}
where $p_i$ is the probability distribution of the target, $i$ is the number of classes, $\bar{p}_i$ is the predicted probability distribution and  $t$ is the target class. 

According to \cite{7410714}, the lower layers of deep architectures encode more discriminative features to capture intra-class variations and provide more detailed
local features and the higher layers of deep architectures
focus on identifiable local semantic concepts. We borrowed this idea and compare the images based on low level features for verification and extract
identification features from higher layers as in our work, we need to focus on intra-class variations for verification purpose and local semantic concepts for identification.

\begin{table*}
\fontsize{7}{7}\selectfont
\begin{tabular}{|p{2.7cm}|p{0.4cm}|p{0.3cm}|p{0.3cm}|p{0.3cm}||p{0.3cm}|p{0.3cm}|p{0.3cm}||p{0.3cm}|p{0.3cm}|p{0.3cm}||p{0.3cm}|p{0.3cm}|p{0.3cm}||p{0.3cm}|p{0.3cm}|p{0.3cm}||p{0.3cm}|p{0.3cm}|p{0.3cm}|}
\hline
\multirow{3}{*}{Method} & \multirow{3}{*}{Type} & \multicolumn{3}{c|}{CUHK01} & %
    \multicolumn{3}{c|}{CUHK03} & \multicolumn{3}{c|}{VIPeR} & \multicolumn{3}{c|}{PRID2011} & \multicolumn{3}{c|}{iLIDS} & \multicolumn{3}{c|}{Market1501} \\
\cline{3-20}
& & R1 & R5 &R10  & R1 & R5 & R10 & R1 & R5 & R10& R1 & R5 & R10& R1 & R5 & R10 & R1 & R5 & R10 \\
\hline
DML \cite{6976727} & ID &- &- &- &- &- &- &28.2 &59.2 &73.4 &17.9 &37.5 &45.9 &- &- &- &- &- &-\\
FPNN \cite{6909421} & ID &27.9 &- &- &20.7 &51.3 &68.7 &- &- &-  &- &- &- &- &- &- &- &- &-\\
DRDC \cite{Ding:2015:DFL:2796563.2796623} &V &- &- &- &- &- &- &40.5 &60.8 &70.4 &- &- &- &52.1 &68.2 &75.1 &- &- &-\\
FLCA \cite{7900000}  & ID &46.8 &71.8 &80.5 &-  &- &- &42.5 &72 &91.7 &- &- &- &- &- &- &- &- &\\
DGD \cite{DBLP:journals/corr/XiaoLOW16} &ID &71.7 &88.6 &92.6 &75.3 &- &- &38.6 &- &-  &64 &- &- &64.6 &- &- &- &- &-\\
Improved Trp \cite{Cheng_2016_CVPR} &V &53.7 &84.3 &91.0 &-  &- &- &47.8 &74.7 &84.8  &22.0 &47.0 &57.0 &60.4 &82.7 &90.7  &- &- &-\\
SIRCIR \cite{7780513} & ID+V &71.8 &- &- &52.2 &- &- &35.8 &- &-  &- &- &- &- &- &-  &- &- &-\\
DRPR \cite{DBLP:journals/corr/ChenGL15} &V &70.9 &92.3 & 96.9 &-  &- &- &38.3 &69.2 &81.3  &- &- &- &- &- &-  &- &- &-\\
PersonNet \cite{DBLP:journals/corr/WuSH16} &V &71.1 &90 & 95 &64.8 &89.4 &94.9 &- &- &-  &- & -&- &- &- &- &-&-&-\\
DLCNN \cite{DBLP:journals/corr/ZhengZY16} & ID+V &- &- &- & 83.4 &97.1 &98.7 &- &- &-  &- &- &- &- &- &- &- &- &-\\
GSCNN \cite{DBLP:journals/corr/VariorHW16} & ID &- &- &- &68.1 &88.1 &94.6 &37.8 &66.9 &77.4  & -&- &- &- &- &-  &65.9 &- &-\\
CAN \cite{DBLP:journals/corr/LiuFQJY16} & ID &- &- &- &72.3 &93.8 &98.4 &- &- &-  &- &- &- &- &- &-  &60.3 &- &-\\
Re-ranking \cite{Re-ranking} &ID &- &- &- &61.6 &- &- &- &- &-  &- &- &- &- &- &-  &77.1 &- &-\\
DCF \cite{Li_2017_CVPR} &ID &- &- &- &74.2 &94.3 &97.6 &- &- &-  &- &- &- &- &- &-  &80.3 &- &-\\
SSM \cite{Bai2017ScalablePR} &V &- &- &- &76.6 &94.6 &98.0 &53.7 &- &91.5  &- &- &- &- &- &-  &82.2 &- &-\\
EDM \cite{Shi2016} & ID &69.4 &- &- &61.3 &- &- &40.9 &- &-  &- &- &- &- &- &-  &- &- &-\\
MTDNet \cite{AAAI1714313} & ID+V &77.5 &95.0 &97.5 &74.7 &95.9 &97.5 &45.9 &71.9 &83.2 &32.0 &51.0 &62.0 &- &- &- &- &- &\\
BTL \cite{DBLP:journals/corr/ChenCZH17} & V &62.6 &83.4 &89.7 &75.5 &95.1 &99.1 &49.0 &73.1 &81.9  &- &- &- &- &- &- &- &- &-\\
P2S \cite{8100017} &V &77.3 &93.5 &96.7 &-  &- &- &- &- &-  &70.7 &95.1 &98.9 &- &- &- &70.7 &- &-\\
Spindle Net \cite{zhao2017spindle} &ID &79.9 &94.4 &97.1 & 88.5 &97.8 &98.6 &53.8 &74.1 &83.2  &67.0 &89.0 &89.0 &66.3 &86.6 &91.8 &76.9 &91.5 &94.6\\
DaRe \cite{DaRe} &V &- &- &- &73.8 &- &- &- &- &-  &- &- &- &- &- &- &90.9 &- &-\\
AACN \cite{Xu2018AttentionAwareCN} &ID &88.1 &96.7 &98.2 &91.4 &98.9 &99.5 &- &- &-  &- &- &- &- &- &- &88.7 &- &-\\
MLFN \cite{98a1e05749b24099a51dcf3c22daefd9} &ID &- &- &- &82.8 &- &- &- &- &-  &- &- &- &- &- &-  &90.0 &- &-\\
DFSN \cite{Amena} &ID+V &83.9 &98.2 &98.9 &85.5 &98.7 &99.8 &68.7 &88.9 &94.6  &75.0 &93.0 &97.0 &- &- &-  &- &- &-\\
CamStyle \cite{zhong2018camera} &ID &- &- &- &- &- &- &- &- &-  &- &- &- &- &- &-  &89.5 &- &-\\
\hline
SC-IMGAN + Imp Quartet &ID+V  &\textbf{95.3}  &\textbf{99.5}  &\textbf{99.8} &\textbf{92.8}  &\textbf{99.8}  &\textbf{99.8} &\textbf{72.8}  &\textbf{95.1}  &\textbf{98.0}  &\textbf{80.7}  &\textbf{96.2}  &\textbf{99.4} &\textbf{86.0}  &\textbf{96.2}  & \textbf{98.3}  &\textbf{94.7} &\textbf{95.7} &\textbf{98.5}\\
\hline
\end{tabular}
\caption{\label{tab:results} Comparison of SC-IMGAN with state-of-the-art methods on CUHK01, CUHK03, VIPeR, PRID2011, iLIDS and Market-1501 datasets. ID, V and ID+V indicate that an identification, a verification or a combination is used. R1, R5 and R10 indicates rank-1, rank-5 and rank-10 identification accuracy respectively.}
\vspace{-4mm}
\end{table*}

\vspace{-2mm}
\subsection{Training the Network}
\vspace{-2mm}
For style-transferred image generation, we use the training data from six person Re-ID databases to train SC-IMGAN. For each training image, we map to each target domain, such that the network learns how to map each image to each domain. We use pytorch and we empirically set $\lambda_1$ o 10, and $\lambda_2$ and $\lambda_3$ to 0.1 (see Equation \ref{eq:5}). The Adam optimizer is used to train SC-IMGAN from scratch with a batch size of 1 and a learning rate of 0.0002. We train for 200 epochs. The learning rate is constant for the first 100 epochs, and then linearly decays towards zero over the next 100 epochs. The styles are transferred among 6 domains which requires 6 generator/discriminator pairs whereas CycleGAN requires 30 generator/discriminator pairs, i.e. 30 training networks. To train for 200 epochs, the proposed SC-IMGAN takes approximately 120 hours whereas CycleGAN takes around 1560 hours to train the generators, on a single GPU. The generator and discriminator architecture of SC-IMGAN is illustrated in Figure \ref{fig:GAN}. The encoder contains three convolutional layers. The output activation is then passed through a series of nine residual blocks, which is expanded by the decoder. 

For person Re-ID, we use a four stream CNN, fine-tuned from an Alexnet \cite{NIPS2012_4824} model pre-trained on ImageNet \cite{imagenet_cvpr09}. We set the learning rate to 0.001 and use a batch size of 128 during training.  $\tau_1$ is set to -1 and  $\tau_2$ to 0.01 in Equation \ref{eq:6}. The training iterations are set to 30,000 and training takes approximately 24 hours. Stochastic gradient descent (SGD) is used to update network parameters. 
\vspace{-2mm}
\section{Experimental Results and Discussions}
\vspace{-2mm}
\subsection{Datasets}
\vspace{-2mm}

For SC-IMGAN, six Re-ID datasets (CUHK01, CUHK03, VIPeR, PRID2011, iLIDS, and Market1501) are used to train the generator network where the training images (training splits are the same as discussed below) are used to generate synthetic images. For each dataset, the generated synthetic images are used to train the Re-ID network alongside real images.

For Re-ID, we use the above mentioned six datasets separately to evaluate the proposed method; i.e. for the CUHK01 dataset, the Re-ID network is trained with the real images from CUHK01 dataset along-with the generated images from CUHK01 in the style of other domains. CUHK01 \cite{6619305} consists of 3884 images of 971 persons taken by different two camera views, each identity has four images. CUHK03 \cite{6909421} consists of 1467 identities, captured by six surveillance cameras from a university campus. VIPeR \cite{1478416} contains 1264 images of 632 identities, captured by different cameras with changes in viewing angles, poses and lighting conditions. PRID2011 \cite{conf/scia/HirzerBRB11} contains 385 and 749 identities in two camera views which are captured from videos. Among 1134 persons, only 200 are common to both camera views. iLIDS \cite{Wang2014PersonRB}, consists of 479 images of 119 identities which are extracted from video images in a busy airport environment. Market-1501 \cite{Market} consists of 12936 training images of 751 identities and 19732 testing images of 750 identities which is close to a real world setting. For all six datasets, we follow the same settings for training and testing as \cite{zhao2017spindle}.
\vspace{-2mm}
\subsection{Comparison with state-of-the-art approaches}
\vspace{-2mm}
We compare the results of our proposed method with state-of-the-art approaches as shown in Table \ref{tab:results}. For CUHK01, the proposed SC-IMGAN with the improved quartet loss achieves 95.3\% rank-1 accuracy whereas the previous state-of-the-art approach achieved 88.1\%, which used an identification task only. For CUHK03 and Market-1501, the proposed method achieves 92.8\% and 94.7\% rank-1 accuracy respectively. The previous best method achieved 91.4\% on CUHK03, though they used additional semantic information such as pose estimation for identification and rely on body part detectors; and 90.9\% on Market-1501 where features from multiple layers are combined to capture both high-resolution and semantic details and adopted the traditional triplet loss. For VIPeR, PRID2011 and iLIDS, we outperform the previous state-of-the-art methods by 4.1\%, 5.7\% and 19.7\% rank-1 accuracy respectively. 
\vspace{-2mm}
\subsection{Cross-domain Evaluation}
\vspace{-2mm}
For person Re-ID, cross domain experiments are more relevant for real world deployments. To evaluate whether the domain gap is reduced by SC-IMGAN, we trained the network on CUHK03 and Market1501 and tested on PRID2011 as summarised in Table \ref{tab:cross} . From Table \ref{tab:cross}, the model is trained with images from CUHK03 transferred to other domains by SC-IMGAN  alongside real images, and achieves significant performance gains when tested on PRID, e.g., a 13.5\% and 10.5\% increase in rank-1 accuracy compared to \cite{Wei2018PersonTG}. Similar improvements can be observed when trained with the Martket1501 dataset transfered to other domains, outperforming state-of-the-art approaches \cite{Wei2018PersonTG, ATN}. The comparison indicates the effectiveness of the proposed SC-IMGAN in a cross-domain setting.

\begin{table}
\fontsize{6.5}{6.5}\selectfont
\begin{center}
\begin{tabular}{|p{2.0cm}|p{0.27cm}|p{0.25cm}|p{0.27cm}|p{0.27cm}||p{0.27cm}|p{0.27cm}|p{0.27cm}|p{0.27cm}|}
\hline
\multirow{3}{*}{Method}  & \multicolumn{4}{c|}{CUHK03 $\rightarrow$ PRID} &   \multicolumn{4}{c|}{Market1501 $\rightarrow$ PRID }\\
\cline{2-9}
& \multicolumn{2}{c|}{cam1/cam2 } & \multicolumn{2}{c|}{cam2/cam1}  & \multicolumn{2}{c|}{cam1/cam2 } & \multicolumn{2}{c|}{cam2/cam1} \\
\cline{2-9}
& R1  & R10 & R1  & R10 & R1  & R10 & R1  & R10 \\
\hline
PTGAN(cam1) \cite{Wei2018PersonTG}&18.0   &43.5 &6.5 &24.0 &17.5 &50.5 &8.5 &28.5\\
PTGAN(cam2) \cite{Wei2018PersonTG}&17.5   &53.0 &22.5 &54.0 &10.0 &31.5 &10.5 &37.5\\
ATNet(cam1) \cite{ATN} &-   &- &- &-  &24.0 &51.5 &21.5 &46.5\\
ATNet(cam2) \cite{ATN} &-   &- &- &-  &15.0 &51.0 &14.0 &41.5\\
\hline
SC-IMGAN (cam1)  &\textbf{31.5}  &\textbf{50.0} &\textbf{20.0} &\textbf{32.5} &\textbf{28.0} &\textbf{56.5} &\textbf{26.0} &\textbf{49.5}\\
SC-IMGAN (cam2)  &\textbf{28.0}  &\textbf{57.5} &\textbf{36.5} &\textbf{60.5} &\textbf{21.5} &\textbf{55.0} &\textbf{20.5} &\textbf{45.5}\\
\hline
\end{tabular}
\caption{\label{tab:cross} Cross-domain performance comparison on PRID2011 dataset trained with CUHK03 and Market1501 dataset. $cam1/cam2$ indicates that $cam1$ of PRID is used as the query set while $cam2$ is the gallery set and vice-versa. R1 and R10 indicates rank-1 and rank-10 identification accuracy respectively.}
\end{center}
\vspace{-7mm}
\end{table}

\vspace{-2mm}
\subsection{Ablation Study}
\vspace{-2mm}
\textbf{Effectiveness of Improved Quartet Loss}

In this section, we investigate the effectiveness of the improved quartet loss. We train the Re-ID model with both the triplet loss and quartet loss to evaluate the effectiveness of improved quartet loss. From Table \ref{tab:ablation study}, we see that the
rank-1 accuracy of $Imp Quartet$ outperforms $Quartet$ by 6.3\%, 3.2\%, 1.6\%, 3.2\%, 1.2\% and 6.1\% for CUHK01, CUHK03, VIPeR, PRID2011, iLIDS and Market-1501 datasets; indicating that considering the distance between the positive pairs within the loss is important. Further, we show that the proposed multi-component image generation helps to improve performance compared to CycleGAN and StarGAN.

\textbf{Effectiveness of Identity Mapping Loss}

\begin{figure*}
\begin{center}
\includegraphics[width=1.0\linewidth]{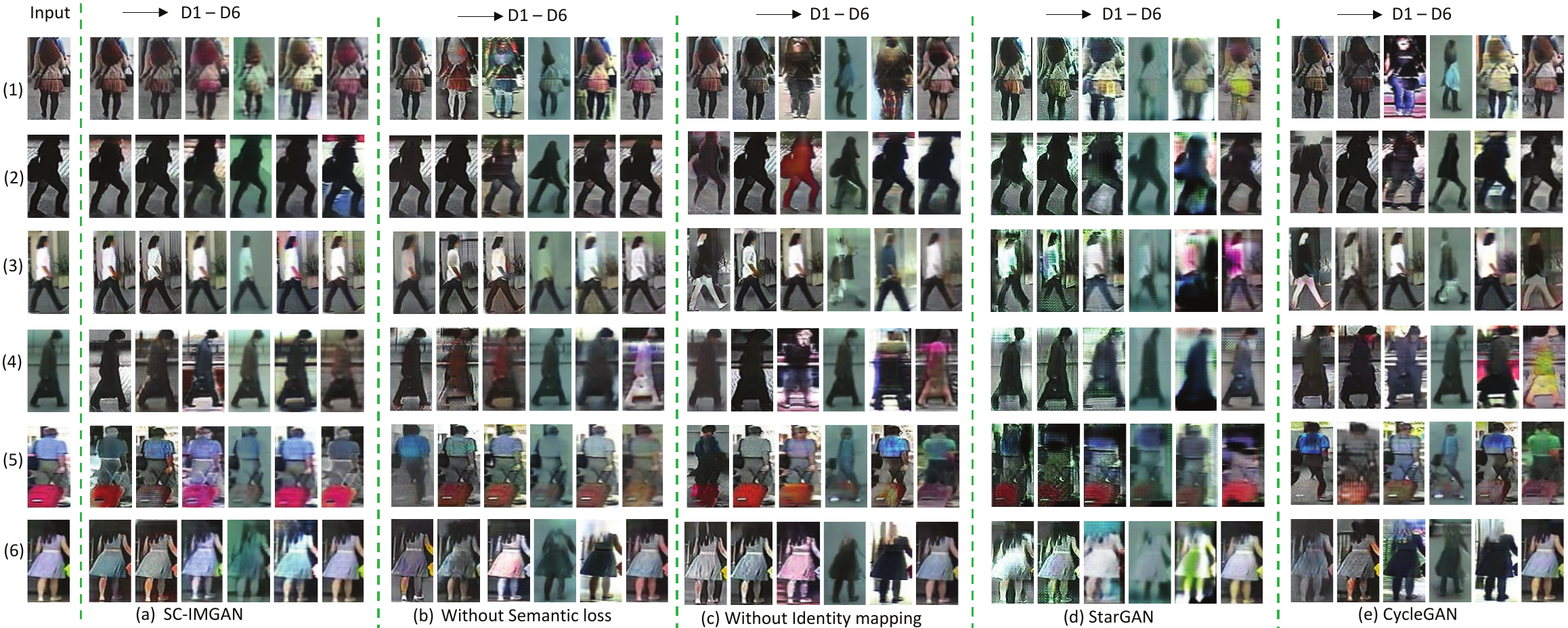}
\end{center}
\vspace{-4mm}
\caption{Examples of style-transferred images in six domains : (a) the samples of the proposed SC-IMGAN (with identity mapping and semantic-consistency loss), (b) samples without semantic-consistency loss, (c) samples without identity mapping loss, (d) StarGAN and (e) CycleGAN. The six rows represent real images from the CUHK01, CUHK03, VIPeR, PRID2011, iLIDS and Market-1501 domains. Inputs from all these domains are transferred to the distributions of CUHK01, CUHK03, VIPeR, PRID2011, iLIDS and Market-1501 respectively, represented as $D1-D6$.}
\label{fig:identity}
\end{figure*}

For person Re-ID, the style transferred image of a person should have the same identity before and after image translation. To this end, our proposed identity mapping multi-component network forces the generator to preserve the identity of the real images such that after style adaptation the identity will be the same. To justify the effectiveness of the identity mapping loss, we compare a multi-component model without the identity mapping loss ($MC-GAN$), with a multi-component GAN with the identity mapping loss but without the semantic-consistency loss ($IMGAN$), as shown in Figure \ref{fig:identity} and Table \ref{tab:ablation study}. From Table \ref{tab:ablation study}, rank-1 performance of IMGAN achieves a 0.9\%, 1.6\%, 0.4\%, 0.7\%, 0.6\% and 1.5\% increase over MC-GAN on CUHK01, CUHK03, VIPeR, PRID2011, iLIDS and Market-1501 datasets. We also observe from Figure \ref{fig:identity} that without the identity loss, it is harder for the model to generate images with the same identity, reducing performance.

\begin{table}
\fontsize{7}{7}\selectfont
\begin{center}
\begin{tabular}{|p{3.0cm}|p{0.28cm}|p{0.3cm}|p{0.3cm}|p{0.3cm}|p{0.3cm}|p{0.3cm}|}
\hline
\multirow{1}{*}{Method} &  \multicolumn{1}{c|}{C1} & 
    \multicolumn{1}{c|}{C3} & \multicolumn{1}{c|}{V} & \multicolumn{1}{c|}{P} & \multicolumn{1}{c|}{L} & \multicolumn{1}{c|}{M} \\
\cline{2-7}
& R1  & R1 & R1 & R1 & R1 & R1\\
\hline
Triplet &79.5   &83.8   &61.0    & 71.0  &71.5   &79.3\\
Quartet &83.9  &85.5  &68.7  &75.0  &82.4 &83.6\\
Imp Quartet &90.2 &88.7 &70.3 &78.2 &83.6 &89.7 \\
CycleGAN + Imp Quartet &90.8  &89.0  &70.9  &78.5  &84.0  &90.2 \\
StarGAN + Imp Quartet &91.4  &89.5  &71.2  &78.8  &84.1  &91.6 \\
MC-GAN + Imp Quartet &93.7  &90.1  &71.5  &79.4  &84.9  &92.1 \\
IMGAN + Imp Quartet &94.6  &91.7 & 71.9  &80.1  & 85.5    &93.6\\
SC-IMGAN + Imp Quartet &\textbf{95.3}  &\textbf{92.8}   &\textbf{72.8}  &\textbf{80.7}  &\textbf{86.0} &\textbf{94.7} \\
\hline
\end{tabular}
\caption{\label{tab:ablation study} Ablation studies on CUHK01, CUHK03, VIPeR, PRID2011, iLIDS and Market-1501 datasets. ``MC-GAN + Imp Quartet'': multi-component image generation network without identity-mapping loss and semantic-consistency loss but with improved Quartet loss. ``IMGAN + Imp Quartet'': with identity mapping loss and improved Quartet loss. ``SC-IMGAN + Imp Quartet'': with semantic-consistency and identity mapping loss and improved Quartet loss. We also compared to StarGAN \cite{8579014} and CycleGAN \cite{CycleGAN2017}. R1 indicates rank-1 identification accuracy.}
\end{center}
\vspace{-7mm}
\end{table}

\begin{figure*}
\begin{center}
\includegraphics[width=1.0\linewidth]{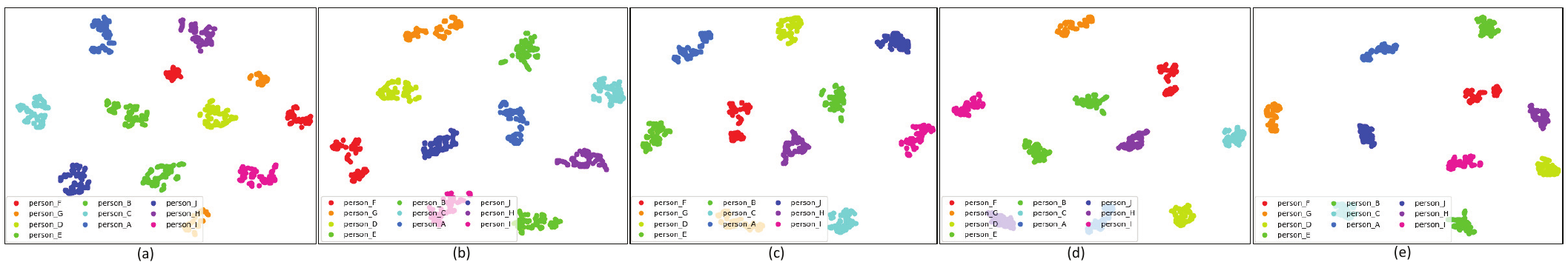}
\end{center}
\vspace{-4mm}
\caption{t-SNE visualizations of the CNN activations for PRID2011 (a) quartet, (b) improved quartet, (c) MC-GAN, (d) IMGAN and (e) SC-IMGAN. The 10 different colours correspond 10 different identities. It can be seen that SC-IMGAN has the tightest grouping of points, indicating that it is best suited to separating the classes.}
\label{fig:Vis}
\vspace{-4mm}
\end{figure*}

\textbf{Effectiveness of Semantic Consistency Loss}

In person Re-ID, domains are vastly different such as between the PRID2011 and Market-1501 datasets. This leads to a performance drop when transforming the images from one domain to others. Thus,
preserving semantic information across domains helps improve translation across domains, and improves performance. To evaluate the effectiveness of the proposed feature-level loss, we evaluate a multi-component model for target domain style adaptation without the semantic-consistency loss ($IMGAN$),  and compare this to the proposed approach (see Figure \ref{fig:identity} and Table \ref{tab:ablation study}). According to Table \ref{tab:ablation study}, removing the the semantic-consistency loss drops performance by 0.7\%, 1.1\%, 0.9\%, 0.6\%, 0.5\% and 1.1\% on the CUHK01, CUHK03, VIPeR, PRID2011, iLIDS and Market-1501 datasets, indicating that the semantic consistency loss leads to a substantial improvement in performance.

From the above, we see that both the semantic consistency loss and identity mapping loss improve performance. It is clear that person Re-ID also benefits from being trained with multiple different styles of a person. Inspecting Figure \ref{fig:identity}, the proposed SC-IMGAN helps preserve the identity and semantics during image domain translation. The proposed model overcomes the limitation of CycleGAN only being able to transfer between two domains by transferring styles among $N$ domains, greatly reducing the computational requirements over CycleGAN. SC-IMGAN also performs better than StarGAN which uses an external code for image translation, and thus can not perform well when domains are vastly different. 

From Table \ref{tab:results} and \ref{tab:ablation study}, it can be observed that the improved quartet loss outperforms \cite{Amena}. While \cite{Amena} used a quartet of images, it did not take into account the distance between the positive pairs and therefore features from the same identity may be separated by a large intra-class distance which leads to a drop in performance. The proposed improved quartet overcomes this problem by not only increasing the inter-class distance with multiple different probe images, but also simultaneously penalising large intra-class distances.

To obtain further insight, a t-SNE visualisation of learned embeddings for the PRID2011 dataset is shown in Figure \ref{fig:Vis}. Here, we have considered 10 classes for clear visualization. From Figure \ref{fig:Vis}, the proposed improved quartet loss optimizes the embedding space such that the data points with the same identity are closer to each other than we see for the quartet loss, for example, the red points are wrongly clustered in Figure \ref{fig:Vis} (a). The effectiveness of the identity mapping loss and semantic loss can also be clearly observed.
\vspace{-3mm}
\section{Conclusion}
\vspace{-3mm}
In this paper, we propose SC-IMGAN, a semantic-consistency and identity mapping multi-component GAN for deep person Re-ID. The SC-IMGAN model learns to transfer styles among six domains to generate new training images. An identity mapping loss ensures that style-transferred images contain the same identity as the original images, and a semantic-consistency loss is also proposed to ensure that encoders extract the same high level features for images belonging to the same identity, regardless of the image domains; improving performance gain during cross-domain translation. The style-transferred images are then used with real images to train the proposed four-stream person Re-ID network. To ensure that we maximise the distance between negative pairs relative to the positive pairs and with respect to multiple different probe images, we propose to use an improved quartet loss with joint verification-identification, which further keeps intra-class features close to each other. Evaluations on 6 popular databases show our technique outperforms current state-of-the-art methods. To emulate a real-world scenario, we performed cross-domain experiments on the proposed methods and the results demonstrate our architecture superior performance in this challenging real world scenario.
\vspace{-4mm}
\section*{Acknowledgements}
\vspace{-3mm}
This research was supported by the Australian Research Council's Linkage Project ``Improving Productivity and Efficiency of Australian Airports'' (140100282). The authors would also like to thank QUT High Performance Computing (HPC) for providing the computational resources for this research.

{\small
\bibliographystyle{ieee}
\bibliography{egbib}
}
\end{document}